\documentclass{article}
\usepackage{spconf,amsmath,graphicx}
\usepackage{subfigure}
\usepackage{float}
\usepackage{gensymb}
\usepackage{ragged2e}
\usepackage{color}
\usepackage{amssymb}

\title{Relative Pose Estimation for Stereo ROLLING SHUTTER Cameras}

\name{Ke Wang, Bin Fan, and Yuchao Dai*\thanks{Yuchao Dai (daiyuchao@nwpu.edu.cn) is the corresponding author.}}
\address{School of Electronics and Information, Northwestern Polytechnical University, China}
\begin{document}
%
\maketitle
\begin{abstract}
In this paper, we present a novel linear algorithm to estimate the 6 DoF relative pose from consecutive frames of stereo rolling shutter (RS) cameras. Our method is derived based on the assumption that stereo cameras undergo motion with constant velocity around the center of the baseline, which needs 9 pairs of correspondences on both left and right consecutive frames.
The stereo RS images enable the recovery of depth maps from the semi-global matching (SGM) algorithm. With the estimated camera motion and depth map, we can correct the RS images to get the undistorted images without any scene structure assumption. Experiments on both simulated points and synthetic RS images demonstrate the effectiveness of our algorithm in relative pose estimation.
\end{abstract}

\section{Introduction}
\label{sec:intro}
With the advantage of low-cost and simplicity in design, more and more rolling-shutter (RS) CMOS cameras \cite{Meingast_Geometric_2005} have been used in various real-world computer vision applications. Different from their global shutter (GS) counterparts that expose all pixels simultaneously, RS cameras capture the pixels of each row (or column) at consecutive times. Therefore, when there is instantaneous motion between the scene and the camera, the captured image will show rolling shutter effects such as skew or wobble, and the level of rolling shutter effect varies with respect to the camera motion and the object motion.

\begin{figure}
\begin{minipage}[b]{0.6\linewidth}
  \centering
  \centerline{\includegraphics[width=7.25cm]{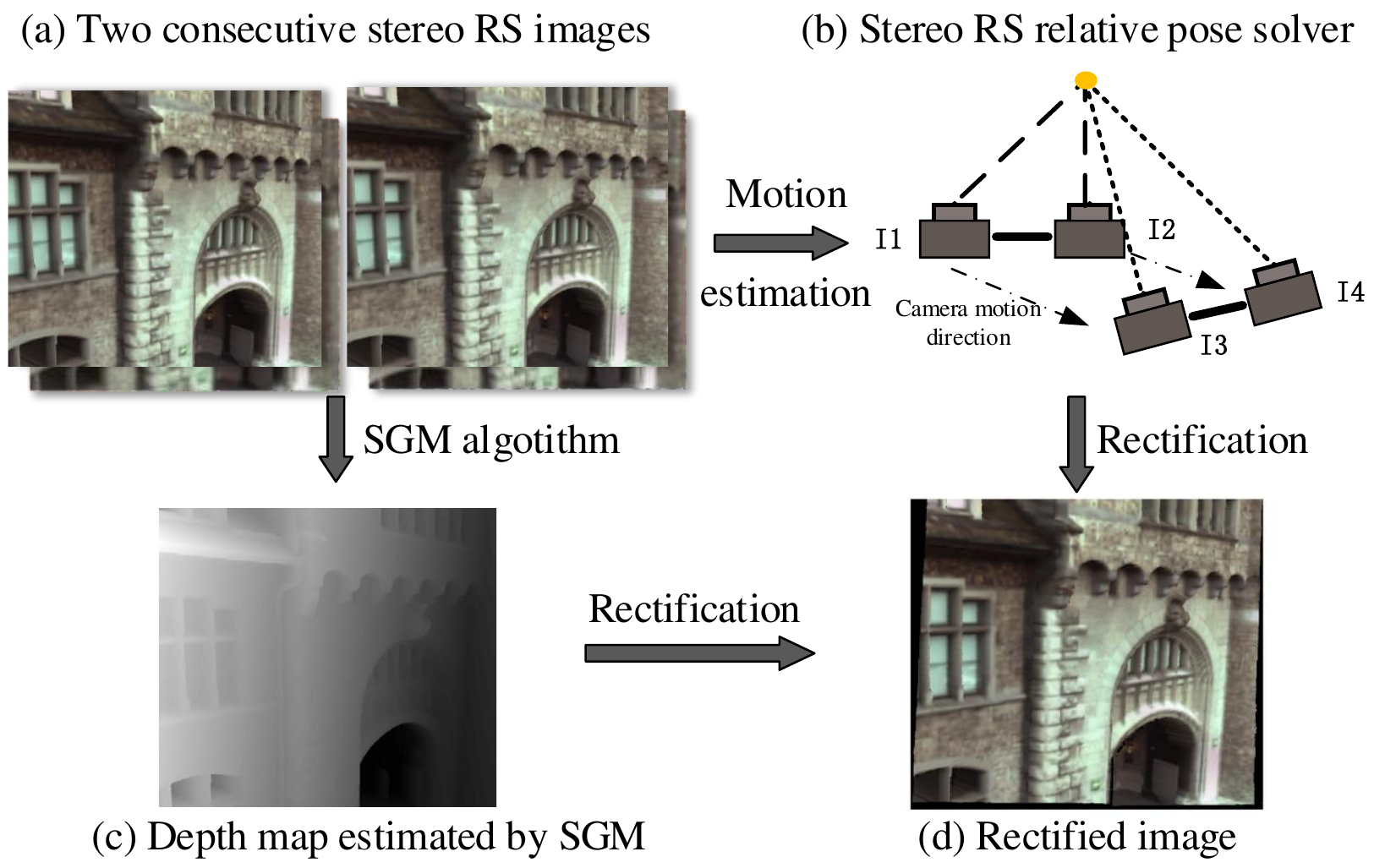}}
\end{minipage}
\centering
\vspace{-0.4cm}
\caption{Overview of our proposed method. From consecutive stereo RS images, we solve for the relative pose via our proposed linear algorithm. With the estimated camera motion and depth map by SGM, we could rectify the RS image.}
\vspace{-0.5cm}
\end{figure}

Currently, most of the geometric algorithms in computer vision are designed under the perspective camera model (\emph{i.e.} GS camera model). However, this model fails to handle the RS images due to the significant difference between RS and GS cameras. Therefore, geometric algorithms for RS cameras have attracted more and more attention. In \cite{Meingast_Geometric_2005}, Meingast \emph{et al} introduced the geometric model for RS cameras. Along this direction, many algorithms have been proposed to solve the corresponding multi-view geometric problems. Purkait \emph{et al}. \cite{Purkait_manhattan_2017} performed pose estimation under the Manhattan world assumption.
Oliver \emph{et al}. \cite{Saurer_RS_stereo_2013} proposed an algorithm based on plane sweeping to recover the depth map and 3D scene for multi-view RS images. Fraundorfer \emph{et al}. \cite{Fraundorfer_Minimal_2010} fused information from external sensors like IMU to estimate the relative motion while Lee \emph{et al} \cite{Lee_Gyroscope_2019} simplified the pose estimation problem by using the gyroscope.
Dai \emph{et al}. \cite{Dai_epipolar_2016} derived a 20-point linear algorithm and a 44-point linear algorithm under different RS models based on generalized epipolar geometry.

It should be noted that all of these works are designed to solve the relative pose of a monocular RS camera. In practise, stereo cameras provide extra benefits over its monocular counterpart, \emph{i.e.,} more accurate motion estimation, depth estimation and the availability of global scale. Existing relative pose estimation methods for stereo cameras such as \cite{Vakhitov_2018_line_point} are all designed for GS cameras.
Surprisingly, no previous attempt has been reported on solving the {\bf 6 DoF relative pose} problem with stereo RS cameras. 
Motivated by relative pose estimation for monocular camera \cite{Dai_epipolar_2016}, we exploit the epipolar geometry to solve the relative pose from consecutive RS frames and present the first relative pose estimation algorithm for stereo RS cameras as shown in Fig.1. 
Specifically, we figure out a linear solver for 6 DoF relative pose from consecutive stereo images under a continuous motion model. To solve the pose, 9 pairs of correspondences from the left and right cameras respectively are needed. The pose of each scanline on each frame can be solved under the assumption of a constant velocity motion model.

The estimated 6 DoF camera motion also enables the correction of the RS effects, which generally requires the underlying scene structures and the camera motions between scanlines or between views. Existing work with monocular cameras depend on strong assumptions to estimate the scene structure. 
Lao \emph{et al}. \cite{Lao_2018_Robust} used at least 4 curves to estimate the camera's rotation speed to eliminate the RS effect. 
Zhuang \emph{et al}.\cite{Zhuang_Rolling_shutter_aware_2017} introduced optical flow based on RS successive frames, which proposed an 8-point algorithm for constant motion, and 9-point algorithm for constant acceleration motion, then they achieved SfM as well as image rectification.
Vasu \emph{et al}. \cite{Vasu_Occlusion_2018} realized that the RS distortion effect is depth-dependent, and proposed an algorithm that can recover 3D scene from a set of RS distorted images. 

Under our stereo RS camera configuration, we are able to estimate the scene structure without assumptions. Specifically, we show that for stereo RS images, using the semi-global matching (SGM) algorithm which is based on stereo GS image can also accurately estimate the depth map, then the depth will be used in the RS image correction to obtain a smoothly corrected image.
Experimental results on both simulated points and synthetic RS images prove the effectiveness of our relative pose algorithm and its application in RS image correction,



\section{METHOD}
\label{sec:format}
In this section, we establish the model of stereo RS cameras under consecutive frames and derive the linear solver.

\subsection{Rolling shutter camera model}
\label{ssec:subhead}
First of all, we will illustrate the differences between the GS model and the RS model in the form of image formation. GS cameras expose all pixels simultaneously, thus the projection can be expressed as:
$\lambda_{i} \mathbf{x}_{i} = \boldsymbol{\rm K}(\boldsymbol{\rm R} \mathbf{X}_{i} + \mathbf{T})$, where $\boldsymbol{\rm R}$ and $\mathbf{T}$ represent the rotation and translation of the camera in the world coordinate respectively, $\boldsymbol{\rm K}$ denotes the intrinsic calibration matrix, $\mathbf{X}_{i}$ is the 3D point which is projected to $\mathbf{x}_{i}=[u_{i},v_{i},1]^{T}$ in image plane with a depth value $\lambda_{i}$.

For RS camera, each scanline will correspond to different poses when the camera is moving during image acquisition. We assume that the camera scans in rows. So both $\boldsymbol{\rm R}$ and $\boldsymbol{\rm T}$ can be described as functions of image row $u_{i}$. The projection process under RS camera model is expressed as:
\begin{align}
\lambda_{i} \mathbf{x}_{i}= \lambda_{i}[u_{i},v_{i},1]^{T}=
\boldsymbol{\rm K}[ \boldsymbol{\rm R}(u_{i}) \mathbf{X}_{i} + \mathbf{T}(u_{i})].
\end{align}
Next, we derive our linear solver of the stereo RS cameras under consecutive frames.

\subsection{Stereo RS algorithm under consecutive images}
\label{ssec:subhead}
As illustrated in Fig.1(b), we use the continuous camera motion model in this paper. 
For clarity, we denote the left and right camera at the first moment and the second moment as ${\rm I}_{1}$, ${\rm I}_{2}$, ${\rm I}_{3}$, ${\rm I}_{4}$ respectively.

We suppose that stereo cameras undergo constant velocity motion among consecutive frames. As shown in Fig.2, we define the total readout time of one frame as $T_{a}$ and the delay time between consecutive frames as $T_{b}$. The stereo cameras are rigidly connected through the baseline, so $T_{a}$ and $T_{b}$ of the left and right cameras are identical. $\{\mathbf{w}_{1},\mathbf{w}_{2},\mathbf{w}_{3},\mathbf{w}_{4}\}\in so(3)$ and $\{\mathbf{d}_{1},\mathbf{d}_{2},\mathbf{d}_{3},\mathbf{d}_{4}\}\in \mathbb{R}^{3}$ represent the rotation and translation velocities of the first row on ${\mathrm{I}_{1}}$, ${\mathrm{I}}_{2}$, ${\mathrm{I}}_{3}$ and ${\mathrm {I}}_{4}$. Furthermore, we assume that the stereo cameras move around the center of the baseline, so we set pose of the center of baseline as ${\mathbf{w}_{0} = \mathbf{0}, \mathbf{d}_{0}=\mathbf{0}}$, and the length of baseline is set as $2 \mathbf{b}$. Therefore, the pose of first row on ${\mathrm{I}}_{1}$ can be represented as ${\mathbf{w}_{1} = \mathbf{0}, \mathbf{d}_{1}=\mathbf{b}}$, and the pose of first row on ${\rm I}_{2}$ is ${\mathbf{w}_{2} = \mathbf{0}, \mathbf{d}_{2}=-\mathbf{b}}$.
\begin{figure}[htb]
\begin{minipage}[b]{0.7\linewidth}
  \centering
  \vspace{-0.3cm}
  \centerline{\includegraphics[width=7.3cm]{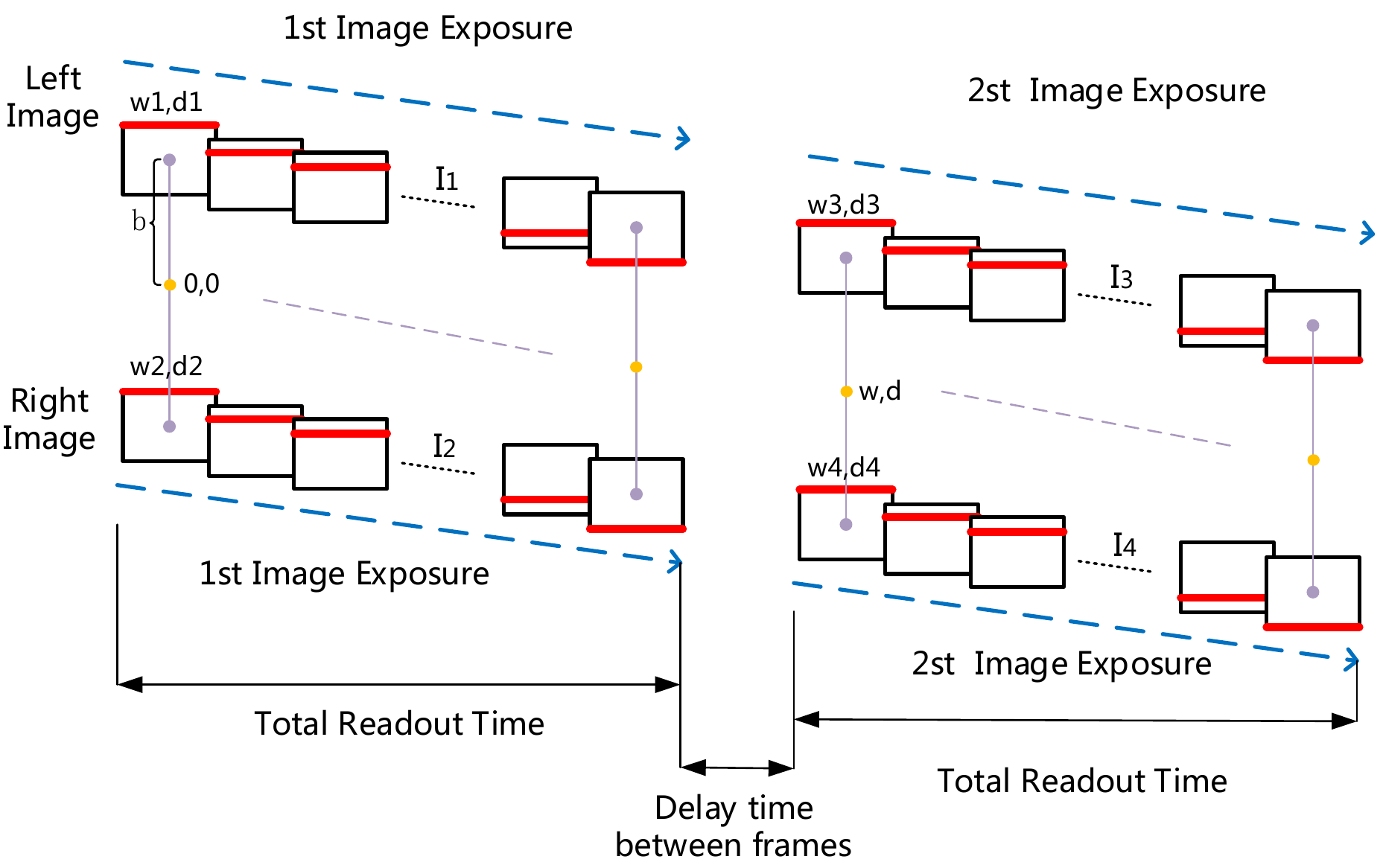}}  
\end{minipage}
\centering
\vspace{-0.5cm}
\caption{Illustration of two consecutive frames of stereo RS cameras.}
\vspace{-0.3cm}
\end{figure}

Suppose that the relative motions between consecutive frames are $\mathbf{w} = [w_{1},w_{2},w_{3}]^{T}$ and $\mathbf{d} = [{\rm d}_{1},{\rm d}_{2},d_{3}]^{T}$, the rotation and translation velocities of each row on each frame can be computed through linear interpolation.
The pose of row $u_{i}$ on ${\rm I}_{1}$ is represented as:
\begin{align}
\mathbf{w}(u_{1}) = hu_{1}\mathbf{w},  \quad   \mathbf{d}(u_{1}) = hu_{1}\mathbf{d} + \mathbf{b},
\end{align}
where $h = \varphi /N$, $\varphi = T_{a} /(T_{a}+T_{b})$ is the readout time ratio, $N$ is the total number of rows per frame. Similar to ${\rm I}_{1}$, the pose of row $u_{3}$ on ${\rm I}_{3}$ is expressed as:
\begin{align}
&\mathbf{w}(u_{3}) = (1+hu_{3})\mathbf{w},  \quad   \mathbf{d}(u_{3}) = (1+hu_{3})\mathbf{d} +\mathbf{b}.
\end{align}
Same as before, the pose of row $u_{2}$ on ${\rm I}_{2}$ and pose of row $u_{4}$ on ${\rm I}_{4}$ are:
\begin{align}
&\mathbf{w}(u_{2}) = hu_{2}\mathbf{w},  \quad   \mathbf{d}(u_{2}) = hu_{2}\mathbf{d} - \mathbf{b},
\end{align}
and
\begin{align}
\mathbf{w}(u_{4}) = (1+hu_{4})\mathbf{w},  \quad   \mathbf{d}(u_{4}) = (1+hu_{4})\mathbf{d} - \mathbf{b}.
\end{align}
Assume that the relative rotation between consecutive frames is small, the mapping between the rotation matrix $\boldsymbol{\rm R}$ and the rotation vector $\boldsymbol{w}$ can be approximated as: $\boldsymbol{\rm R}  \simeq {\rm exp}(\mathbf{w}) =\boldsymbol{\rm I}+[\mathbf{w}]_{\times}$. Therefore, we can develop the essential matrix between ${\rm I_{1}}$ and ${\rm I_{3}}$ as:
\begin{subequations}
\begin{align}
&\boldsymbol{\rm E}_{u_{1},u_{3}} = \left[\mathbf{t}_{u_{1},u_{3}}\right]_{\times} \boldsymbol{\rm R}_{u_{1},u_{3}},\\
&\boldsymbol{\rm R}_{u_{1},u_{3}} = \boldsymbol{\rm R}_{u_{3}}\boldsymbol{\rm R}_{u_{1}}^{T}= \boldsymbol{\rm I}+(1+hu_{3}-hu_{1})[\mathbf{w}]_{\times}, \\
&\mathbf{T}_{u_{1},u_{3}} = \mathbf{d}(u_{3})- \boldsymbol{\rm R}_{u_{1},u_{3}} \mathbf{d}(u_{1}).
\end{align}
\end{subequations}

\noindent Consequently, we can reorganize the formulation to get:
\begin{align}
&\boldsymbol{\rm E}_{u_{1},u_{3}} = a_{1}[\mathbf{d}]_{\times}-a_{1}[[\mathbf{w}]_{\times} \mathbf{b}]_{\times}+{{a_{1}}^{2}}[\mathbf{d}]_{\times}[\mathbf{w}]_{\times}- 
\\
\notag
&a_{1}k_{1}[[\mathbf{w}]_{\times}\mathbf{d}]_{\times}-{{a_{1}}^{2}}[[\mathbf{w}]_{\times}\mathbf{b}]_{\times}[\mathbf{w}]_{\times}-{{a_{1}}^{2}}k_{1}[[\mathbf{\{w}]_{\times}\mathbf{d}]_{\times}[\boldsymbol{w}]_{\times}.
\end{align}
where $a_{1} = k_{3}-k_{1}$,$k_{1} = hu_{1}$, $k_{3}=1+hu_{3}$. Similarly, we can also sort out the essential matrix between ${\rm I_{2}}$ and ${\rm I_{4}}$ as:
\begin{align}
&\boldsymbol{\rm E}_{u_{2},u_{4}} = a_{2}[\mathbf{d}]_{\times}+a_{2}[[\mathbf{w}]_{\times}\mathbf{b}]_{\times}+{{a_{2}}^{2}}[\mathbf{d}]_{\times}[\mathbf{w}]_{\times}-
\\
\notag
&a_{2}k_{2}[[\mathbf{w}]_{\times}\mathbf{d}]_{\times}+{{a_{2}}^{2}}[[\mathbf{w}]_{\times}\mathbf{b}]_{\times}[\mathbf{w}]_{\times}-{{a_{2}}^{2}}k_{2}[[\mathbf{w}]_{\times}\mathbf{d}]_{\times}[\boldsymbol{w}]_{\times},
\end{align}
where $a_{2} = k_{4}-k_{2}$,$k_{2}=hu_{2}$, $k_{4}=1+hu_{4}$. The epipolar constraints are expressed as:
\begin{subequations}
\begin{align}
&[u_{3},v_{3},1]\boldsymbol{\rm E}_{u_{1},u_{3}}[u_{1},v_{1},1]^{{T}}=0,\\
&[u_{4},v_{4},1]\boldsymbol{\rm E}_{u_{2},u_{4}}[u_{2},v_{2},1]^{{T}}=0.
\end{align}
\end{subequations}
By formulating Eq.(9a) and (9b) into the form of a linear equation, we can get $\boldsymbol{\rm A}_{1} \mathbf{X}=0$ and $\boldsymbol{\rm A}_{2}\mathbf{X}=0$ respectively. Combining them together yields a linear equation:
\begin{align}
\left[\begin{array}{c} \boldsymbol{\rm A}_{1}\\ \boldsymbol{\rm A}_{2}\end{array}\right] \mathbf{X}=\boldsymbol{\rm A} \mathbf{X}=0,
\end{align}
where $\mathbf{X} =[d_{1},d_{2},d_{3},w_{1},w_{3},E_{11},E_{12},E_{13},E_{21},E_{22},E_{23},\\
E_{31},E_{32},E_{33},w_{1}w_{1},w_{1}w_{2},w_{1}w_{3},w_{2}w_{3},w_{3}w_{3},S_{1},S_{2},S_{3},\\
S_{4},S_{5},S_{6},S_{7}.S_{8}]^{T}$ in which each of parameter is composed by $\mathbf{w}$ and $\mathbf{d}$, and $\boldsymbol{\rm A}_{1},\boldsymbol{\rm A}_{2}$ are matrices consisted of known parameter $k_{1},k_{2},k_{3},k_{4},b$ and coordinates of correspondences points. In Eq.(10), $\mathbf{X}$ has 27 unknowns. However, we notice that there are higher-order terms in $\boldsymbol{\rm A}$, these terms will not affect the results so we omit them. Then the vector $\mathbf{X}$ contain 19 parameters, thus a linear 18-point solver must exist to solve this formulation, \emph{i.e.} 9 pairs of correspondences for ${\rm I}_{1}$ and ${\rm I}_{3}$ and 9 pairs of correspondences for ${\rm I}_{2}$ and ${\rm I}_{4}$. Next, the right singular vector corresponding to the smallest singular value of $\boldsymbol{\rm A}$ is the solution of Eq.(10). Then, $\mathbf{w}$ and $\mathbf{d}$ can be extracted from $\mathbf{X}$.

\subsection{Rolling Shutter Correction}
\label{ssec:subhead}
In this section, we employ the Semi-Global Matching (SGM) algorithm \cite{Hirschmuller_SGM_2007} to estimate the depth map from the stereo image pair, and combine the camera pose of each row obtained in the previous section, the correction of the RS images can be completed. First, we use the pose and depth of each pixel in the RS image to back-project each pixel into 3D space, and then reproject 3D points to the image plane corresponding to the pose of the first row as follow:
\begin{subequations}
\begin{align}
&\mathbf{X}_{i} = \boldsymbol{\rm R}_{u_{i}}^{T}[\lambda_{i} {\boldsymbol{\rm K}}^{-1} {[u_{i},v_{i},1]}^{T} - \mathbf{T}_{u_{i}}],\\
&{\mathbf{x}_{i}}^{'}= \boldsymbol{\rm K}[ \boldsymbol{\rm R}_{0} \mathbf{X}_{i} + \mathbf{T}_{0}].
\end{align}
\end{subequations}
We figure out that for stereo RS images, a relatively accurate depth map can be obtained by the SGM algorithm, the depth can be further applied to the correction to get a smoothly undistorted image. Note that the stereo RS cameras enable both 6 DoF camera motion estimation and accurate depth map estimation, which  results in our improved RS correction.

\section{EXPERIMENTS}
\label{sec:pagestyle}

In this section, we evaluate our algorithm on simulated points and synthetic RS images. Note that $e_{T}={\rm {acos}}(\mathbf{d}_{est}^{T} \mathbf{d}_{gt}/(\Arrowvert \mathbf{d}_{gt} \Arrowvert \\ \Arrowvert \mathbf{d}_{est} \Arrowvert ))$ and $e_{R}={\rm {acos}}(({\rm Tr}(
\boldsymbol{\rm R}_{est}\boldsymbol{\rm R}_{gt}^{T})-1)/2)$ are used to measure the translation and rotation errors, respectively.

\subsection{Simulation Experiments}
\label{ssec:subhead}
To verify the effectiveness of our proposed algorithm, we simulate matching points that satisfy the RS projection model. In the simulation experiments, we set the size of the image as $900\times900$ with an 810 focal length. Then, we give the value of relative translation $\mathbf{d}$ and relative rotation $\mathbf{w}$. The baseline length $\mathbf{b}$ and the readout time ratio $\varphi$ are also set. When all parameters are known, the correspondences between ${\rm I}_{1}$ and ${\rm I}_{2}$ and correspondences between ${\rm I}_{3}$ and ${\rm I}_{4}$ can be generated. To ensure the creditability, each experimental value is the mean value of 300 repetitions. Moreover, in order to check the robustness and practicability of the algorithm, we do not use RANSAC or nonlinear optimization in the simulation experiments, and experimental results are compared with the experimental results of the GS algorithm \cite{Hartley_Denfense_1997}.

\noindent\textbf{Accuracy versus noise level.} First, we add different levels of random Gaussian noise to the image coordinates of correspondences to evaluate the robustness of our algorithm. Statistical results are shown in Fig.3, illustrating that although the estimation error increases with the increasing noise level, the overall increase is tiny.
\begin{figure*}
\centering
\includegraphics[width=3.7cm]{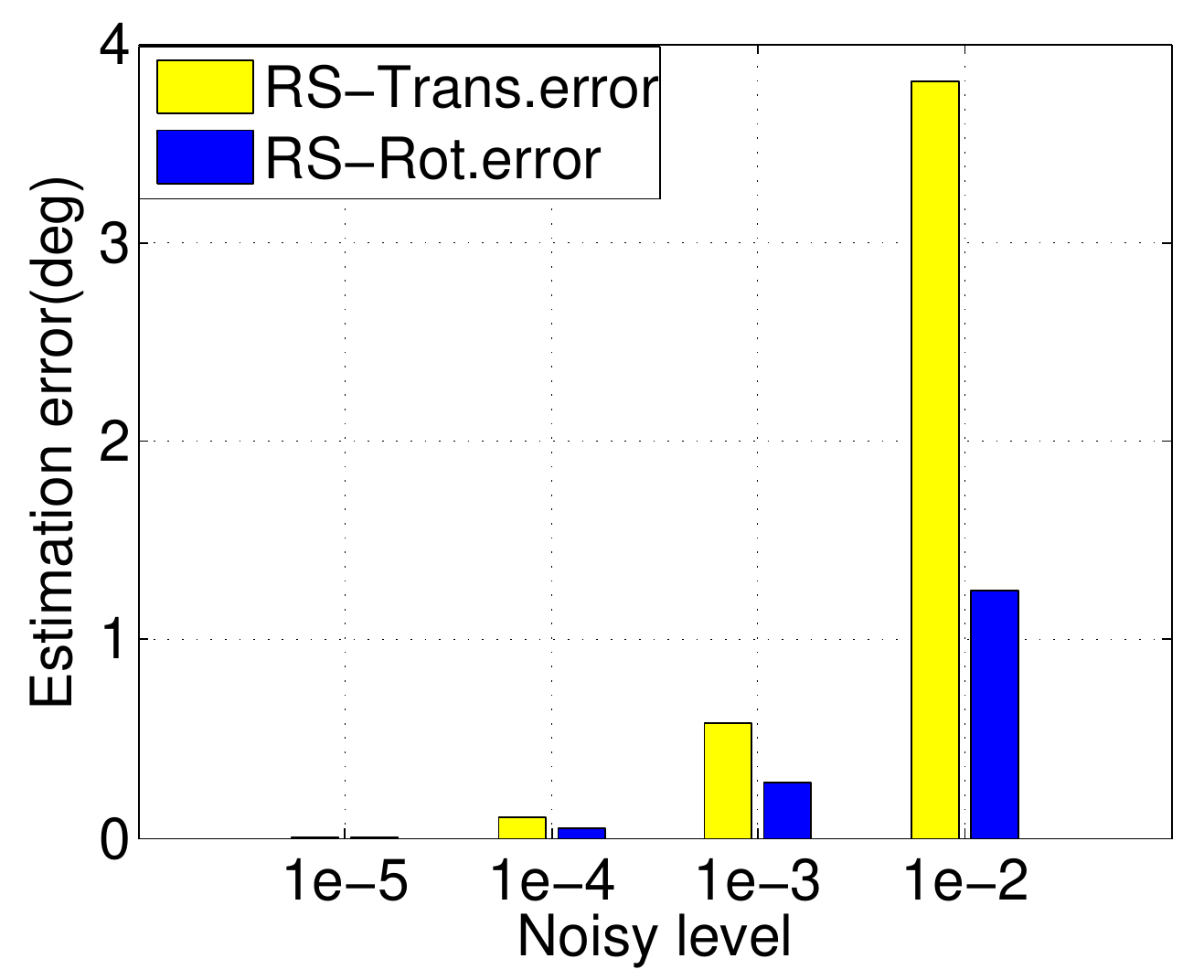}
\hspace{0.1ex}
\includegraphics[width=4cm]{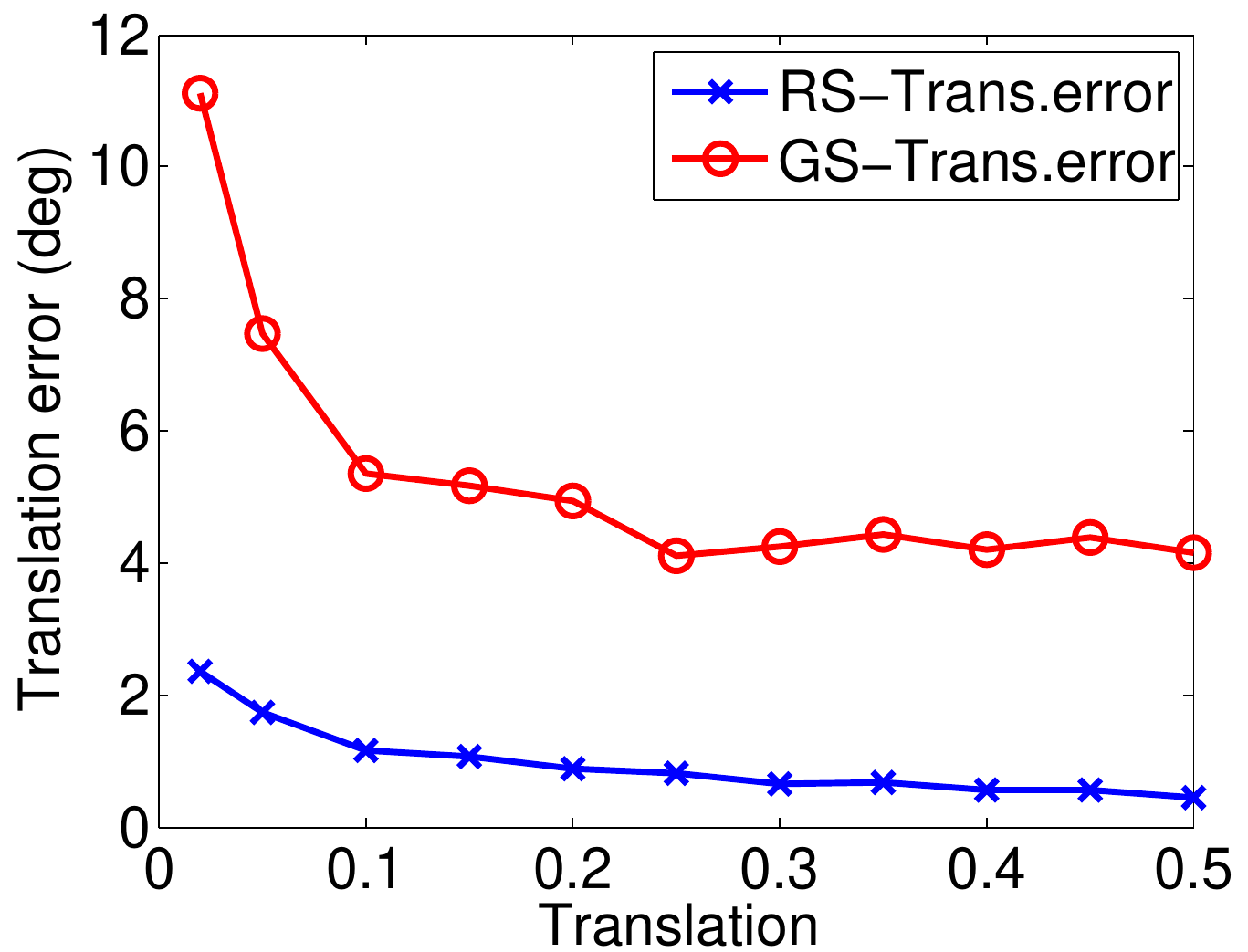}
\hspace{0.1ex}	
\includegraphics[width=3.9cm]{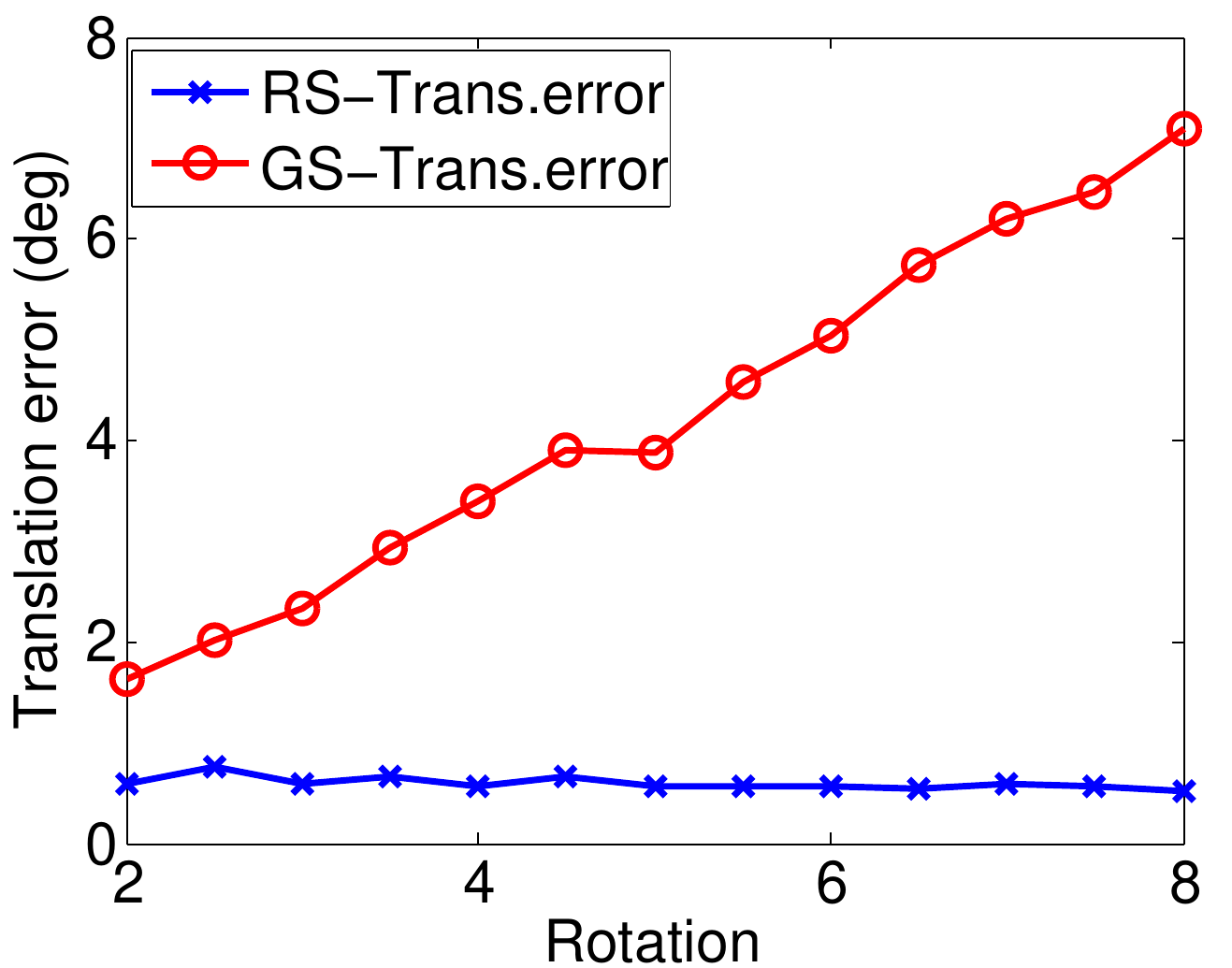}
\hspace{0.1ex}									
\includegraphics[width=3.9cm]{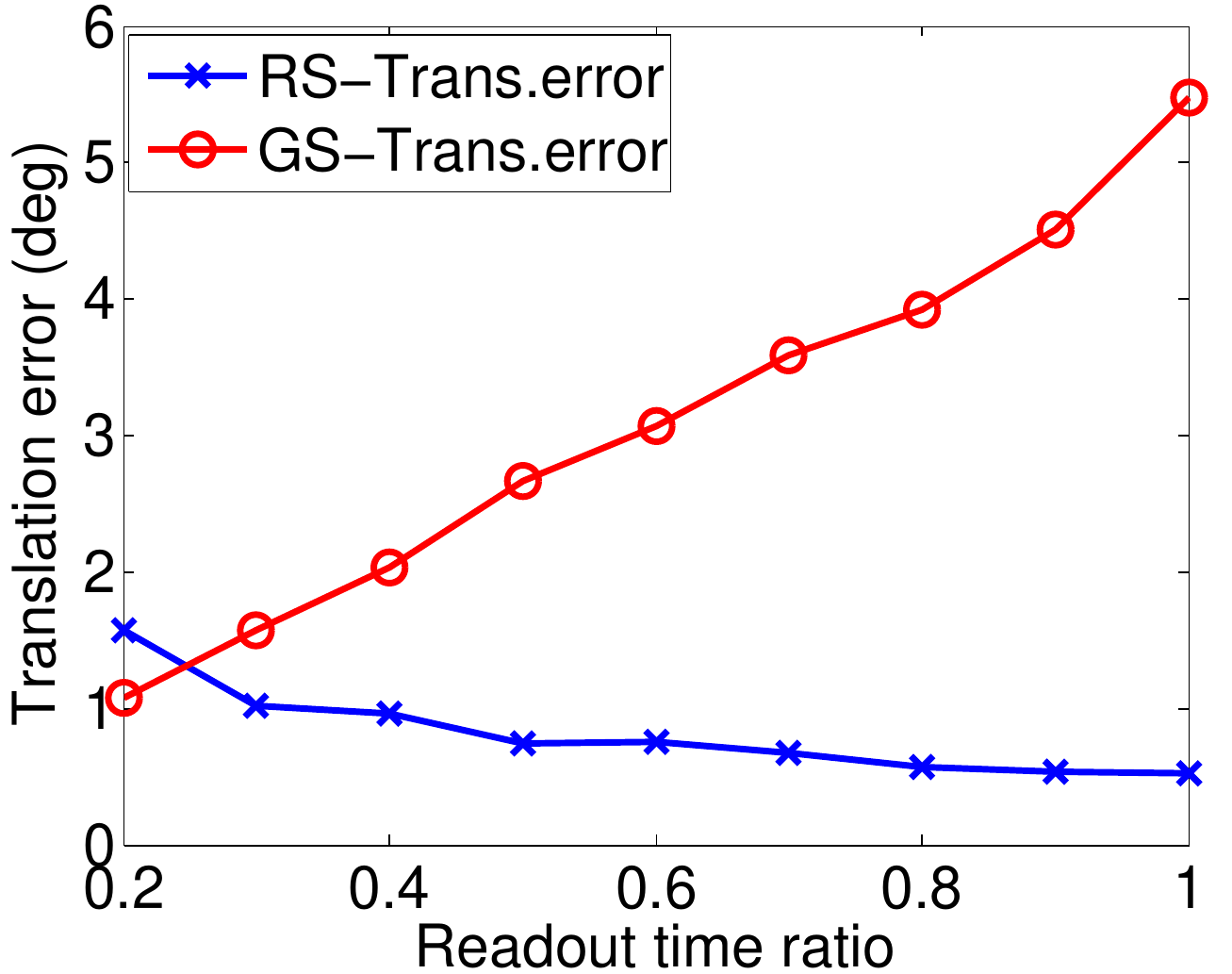}

\vspace{-0.1cm}
\setlength{\abovecaptionskip}{0.cm}
\setlength{\belowcaptionskip}{-0.2cm}
\hspace{-0.8ex}
\centering
\includegraphics[width=3.9cm]{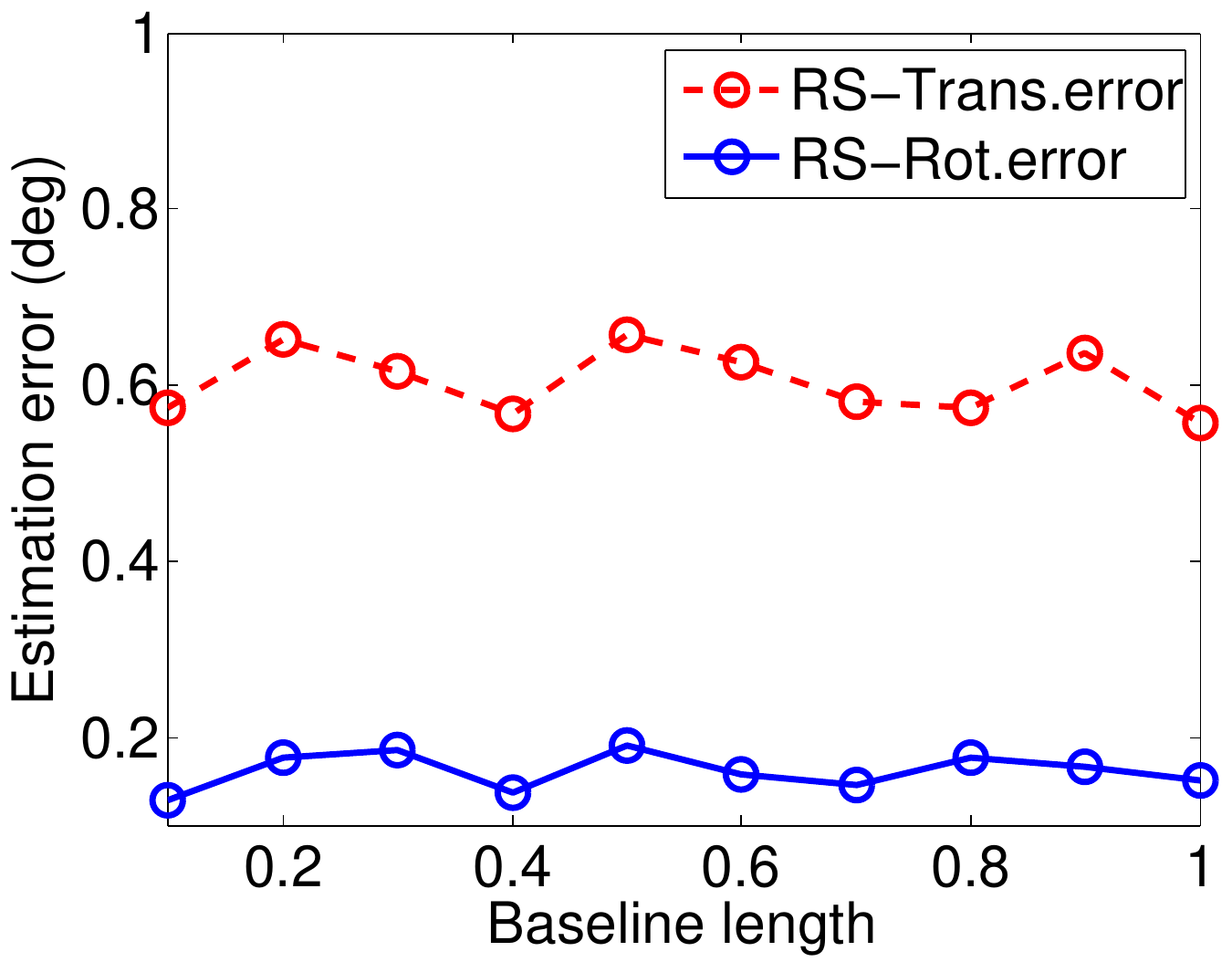}
\hspace{-0ex}
\includegraphics[width=4cm]{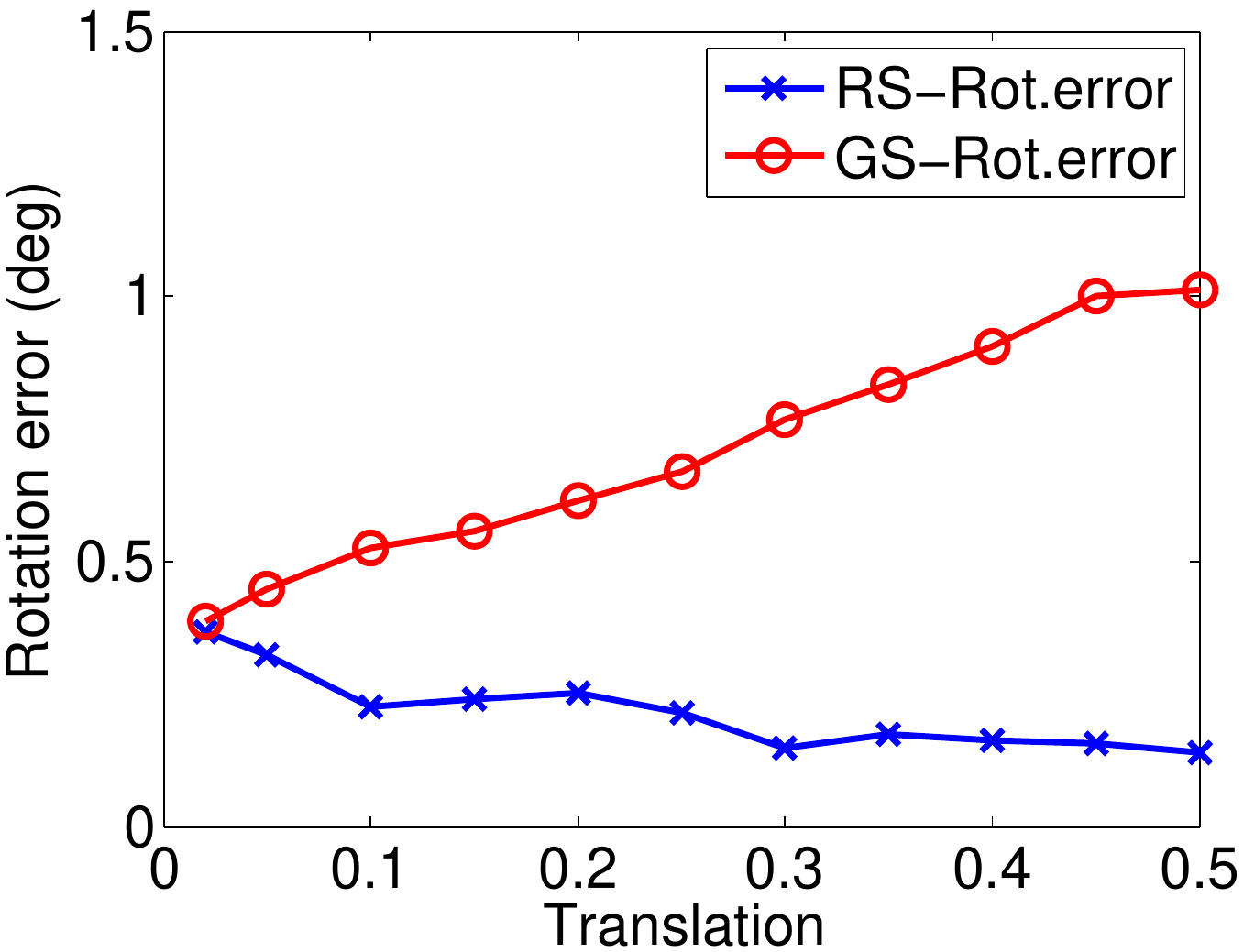}
\hspace{-0.5ex}			
\includegraphics[width=4cm]{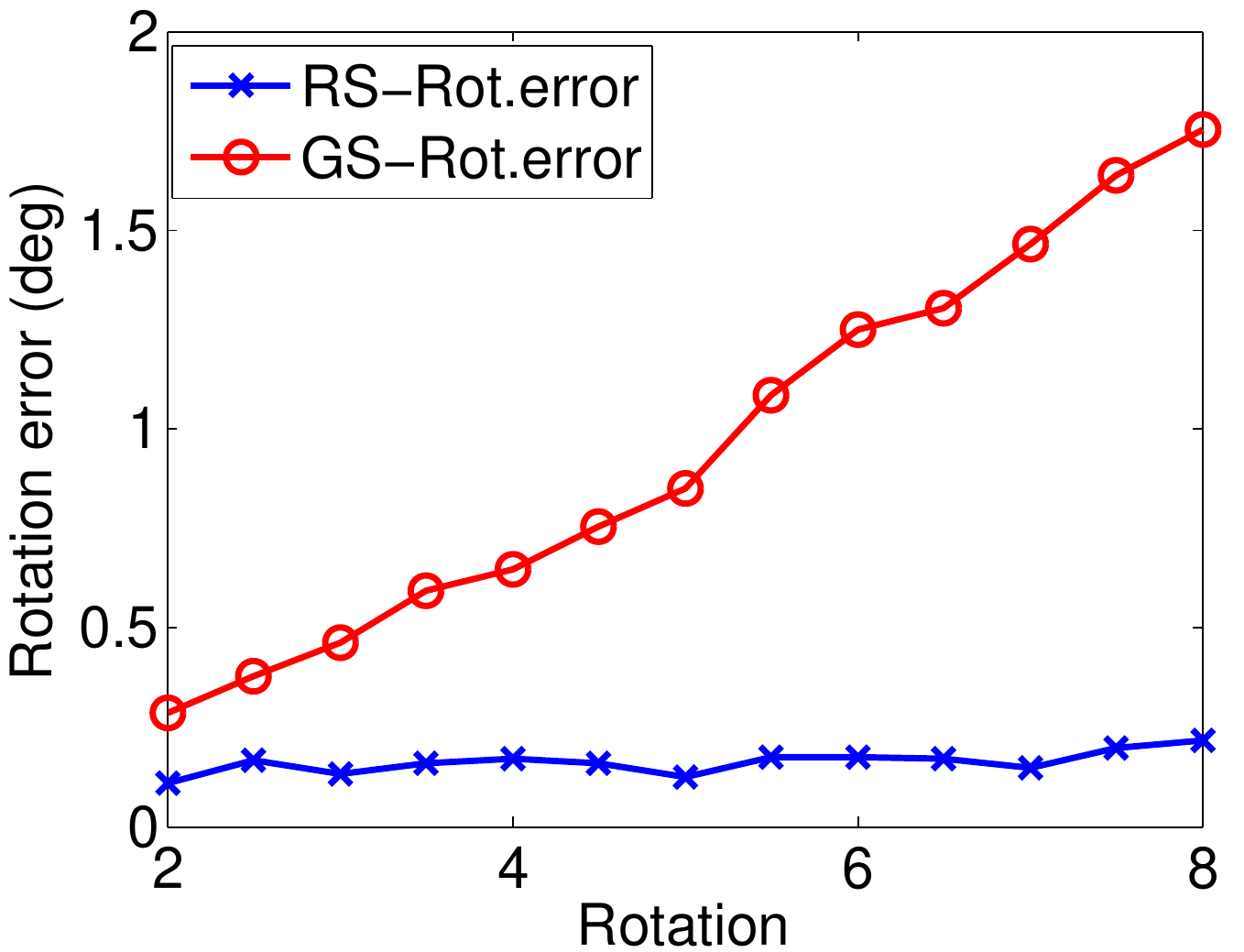}
\hspace{-0.5ex}
\includegraphics[width=4cm]{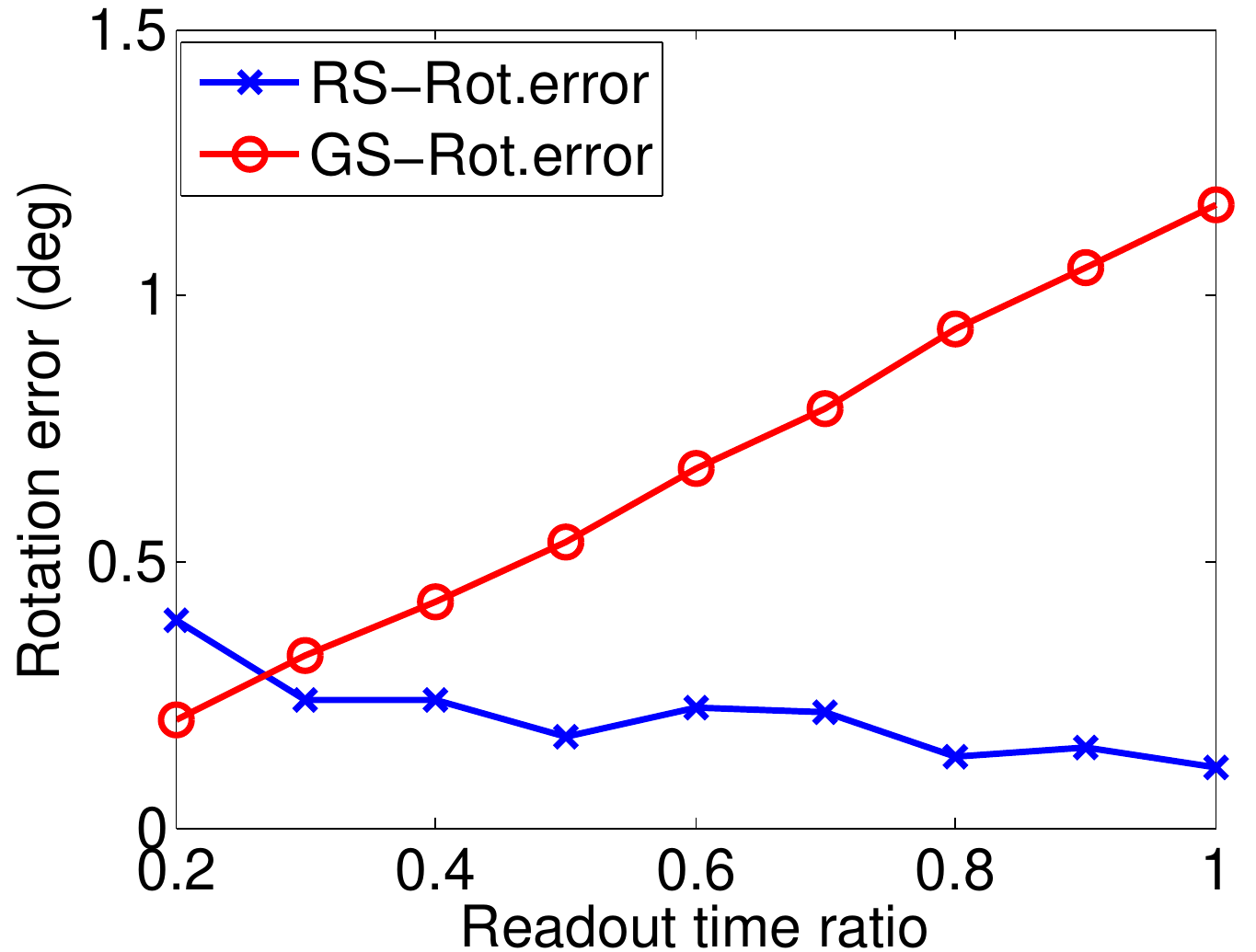}
\vspace{-0.3cm}
\caption{Quantitative evaluation for our stereo RS algorithm on simulation experiments. Results on the first column show the effect of our algorithm under different noise levels and the results with different length of baseline. The second and third columns test the estimation accuracy at different translation and rotation velocity respectively. And the last columns evaluate the effect of different readout time ratio.}
\vspace{-0.5cm}
\end{figure*}

\noindent\textbf{Accuracy versus RS velocity.} Then, we test how RS velocity will affect the accuracy of our algorithm. For a constant Gaussian noise level of $10^{-3}$, we test the performance of the increasing rotation and increasing translation respectively as shown in the second and third column in Fig.3.
It can be observed that our RS algorithm always achieves higher accuracy than the GS algorithm not only in translation estimation but also in rotation estimation with the increasing velocity. Our RS algorithm can maintain more stable and accurate results.

\noindent\textbf{Accuracy versus readout time radio.} To test and observe how the readout time ratio affects the estimation result, we vary the value of readout time ratio $\varphi$ from 0.2 to 1 according to \cite{Im_High_2015}. Simultaneously, the value of $\mathbf{d}$ and $\mathbf{w}$ are kept constant with a fixed Gaussian noise level of $10^{-3}$. Results are shown in the last column in Fig.3, the estimation error of both the GS algorithm and RS algorithm increase with the increase of $\varphi$, but GS gets higher errors. The reason for this result can be interpreted as that the RS effect becomes more evident with increasing $\varphi$.

\noindent\textbf{Accuracy versus baseline length.} Finally, we investigate how the length of the baseline can affect performances. We keep the value of $\mathbf{d}$ and $\mathbf{w}$ fixed, then increase the baseline from 0.1 to 1. As illustrated in the first column in Fig.3, the length of the baseline will not have a significant impact on the performance of results.

\begin{figure}
\vspace{-0cm}
\centering
\hspace{-1.5ex}
\subfigure[Original RS image]{
\begin{minipage}{2.7cm}
\centering
\includegraphics[width=2.45cm]{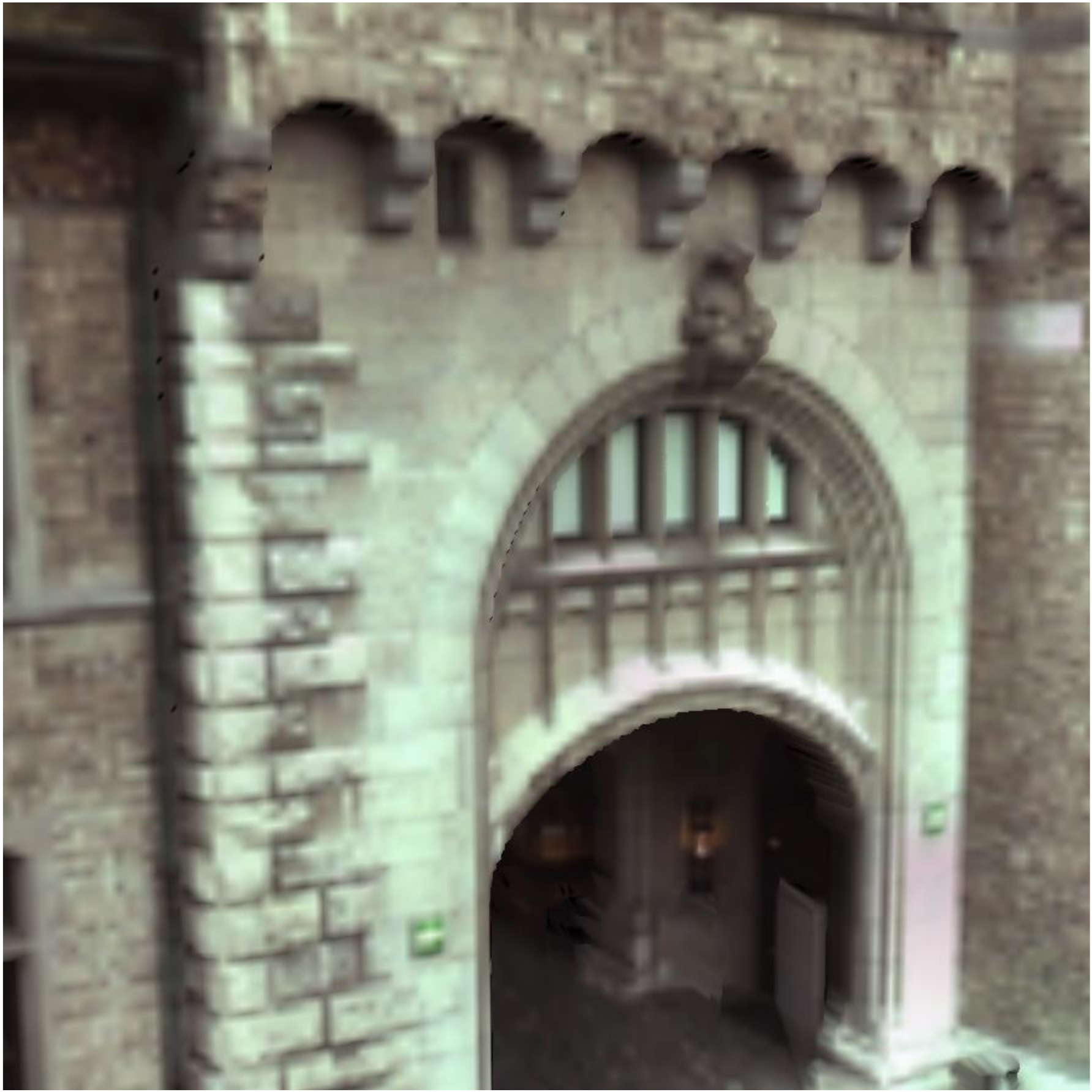}
\end{minipage}%
}%
\subfigure[Corrected image]{
\begin{minipage}{2.7cm}
\centering
\includegraphics[width=2.5cm]{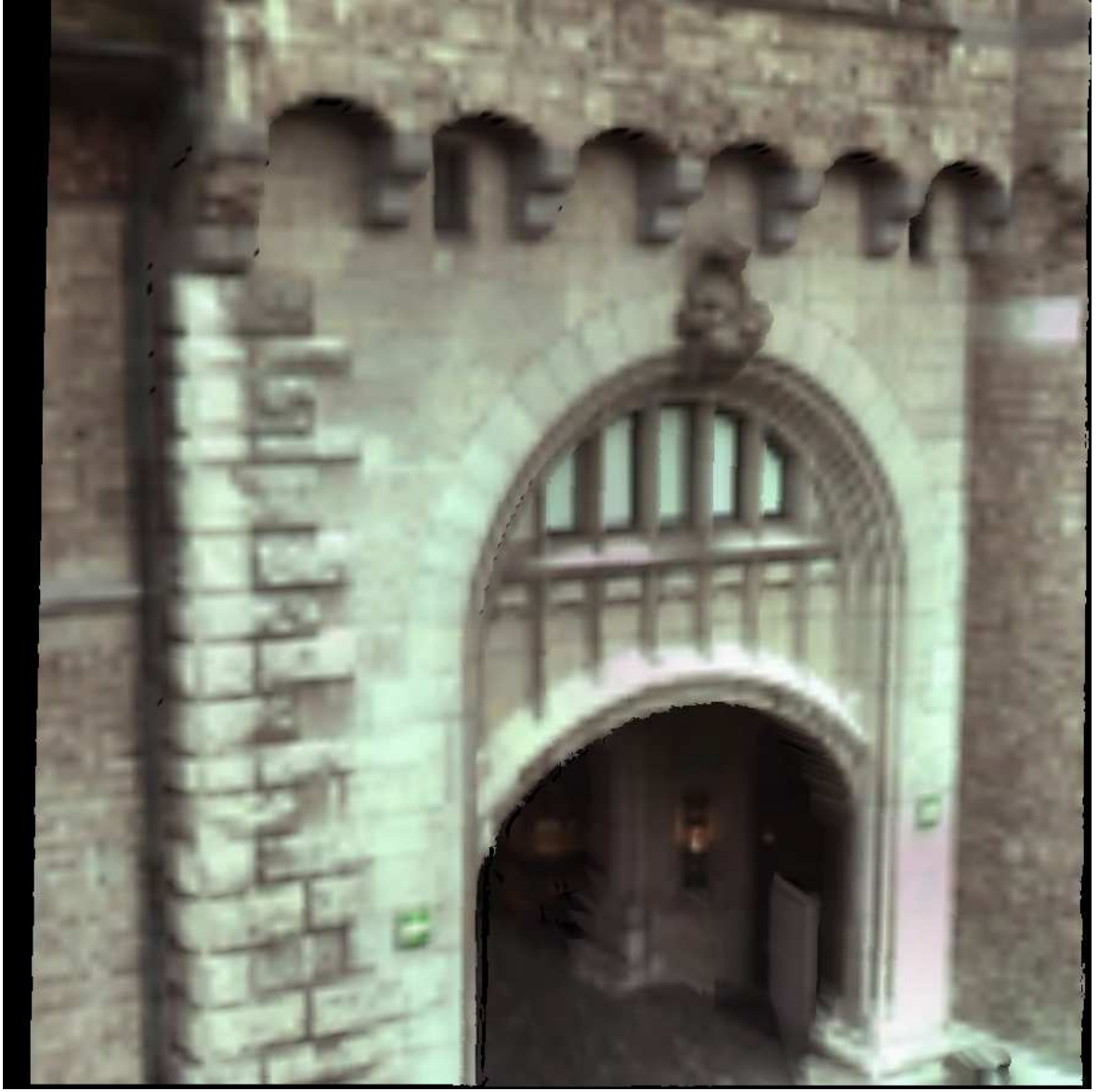}
\end{minipage}%
}%
\subfigure[RS depth truth]{
\begin{minipage}{2.7cm}
\centering
\includegraphics[width=2.5cm]{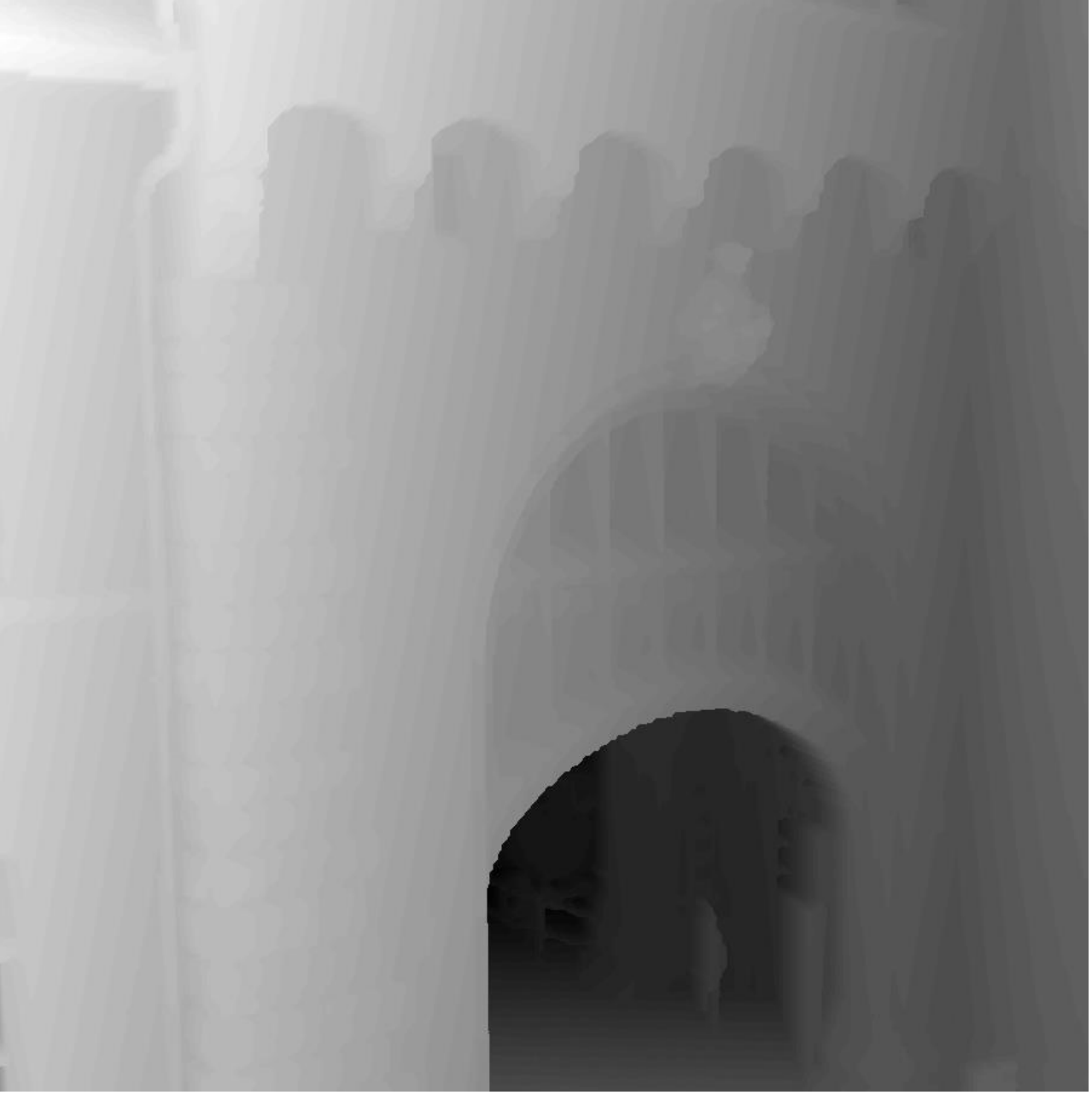} 
\end{minipage}}

\noindent\subfigure[Original RS+GS]{ 
\begin{minipage}{2.7cm}
\centering
\includegraphics[width=2.5cm]{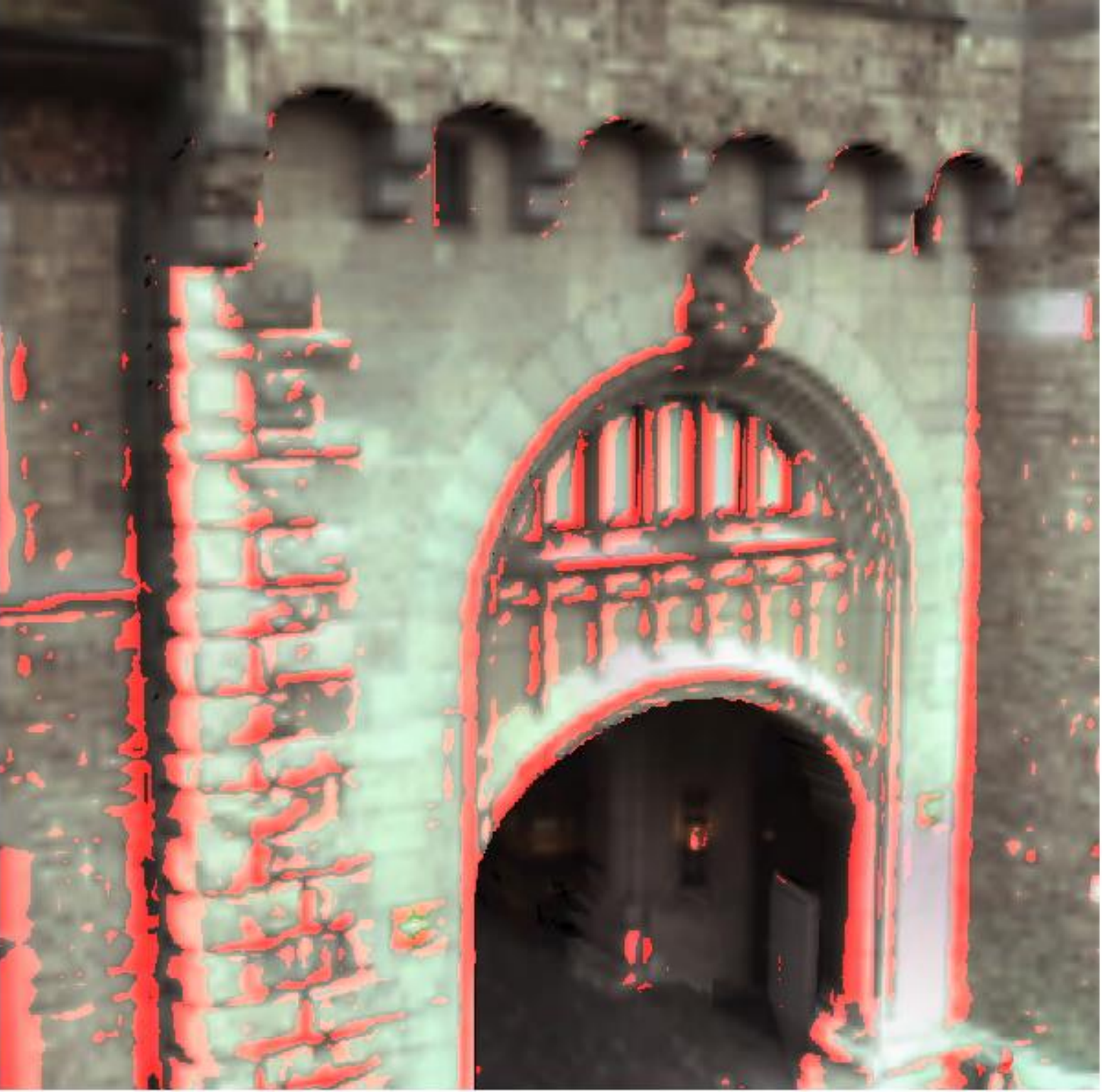}
\end{minipage}}%
\subfigure[Corrected+GS]{
\begin{minipage}{2.7cm}
\centering
\includegraphics[width=2.5cm]{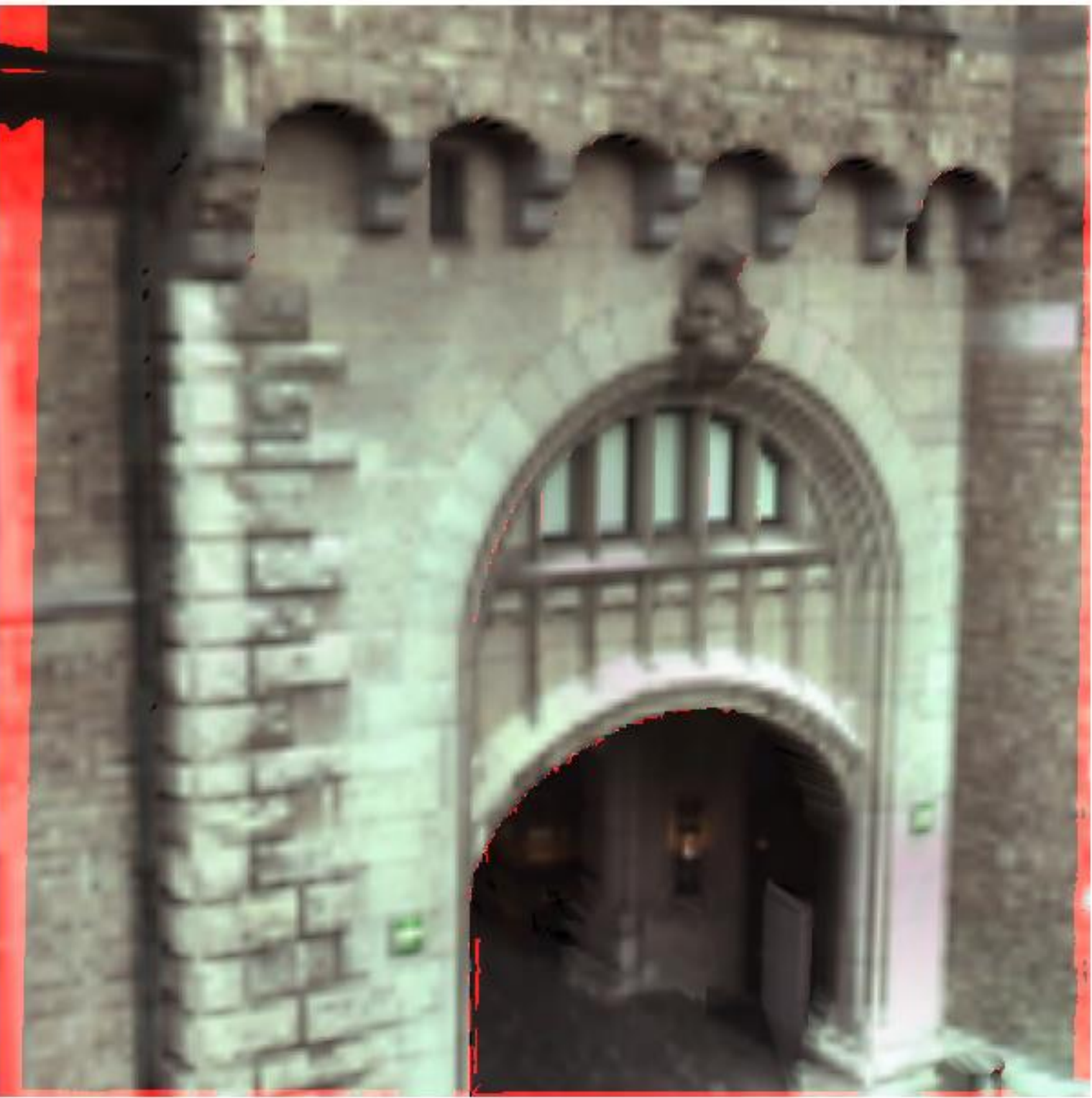}
\end{minipage}}
\subfigure[Depth error]{
\begin{minipage}{2.7cm}
\centering
\includegraphics[width=3.1cm]{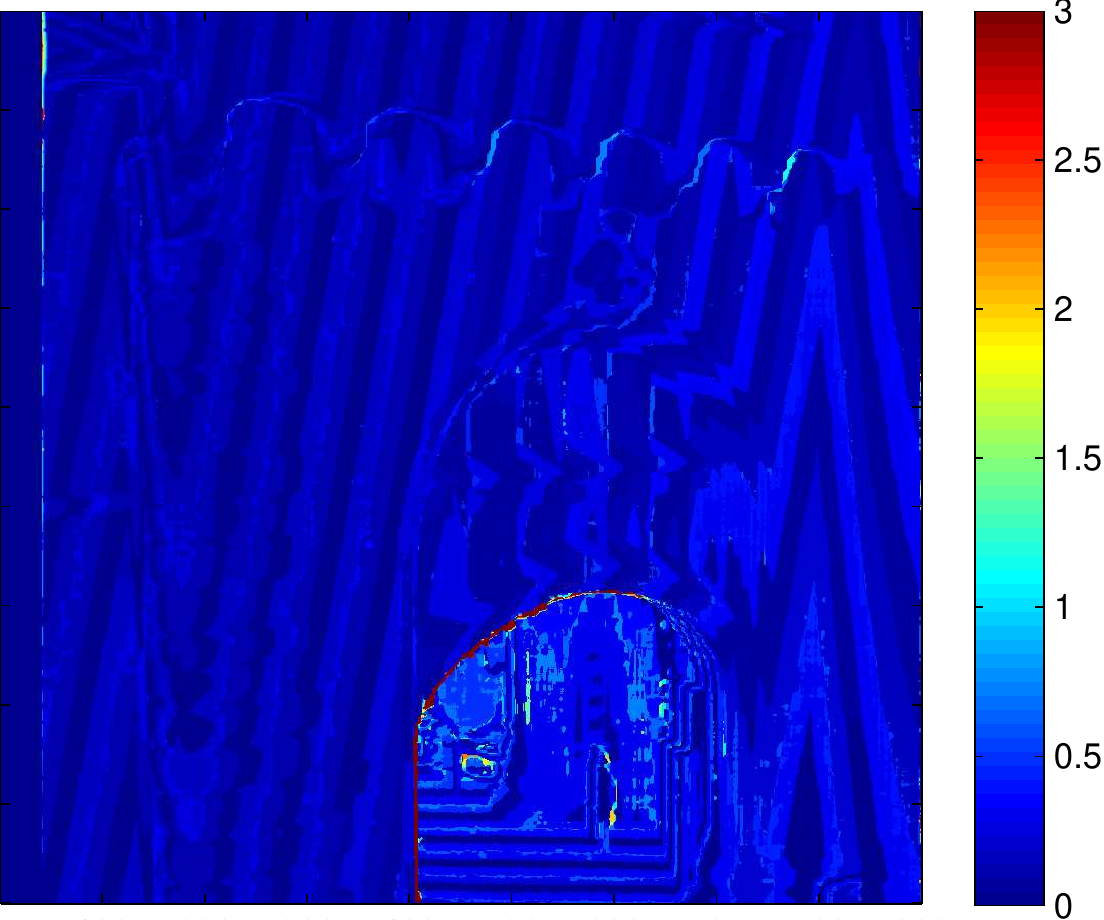} 
\end{minipage}}
\centering
\vspace{-0.3cm}
\caption{Qualitative experiment on synthetic RS images. The experiments exhibit the results of RS image correction and the depth error estimated by SGM algorithm.}
\vspace{-0.5cm}
\end{figure}
\subsection{Experiments on synthetic RS images}
\label{ssec:subhead}
To demonstrate the qualitative results of our relative pose solver, we generate RS images by the simulator and 3D models as in \cite{Zhuang_Rolling_shutter_aware_2017}. We first set the ground truth of all parameters such as $\mathbf{d}$, $\mathbf{w}$ and $\varphi$, then we can calculate the pose of each row on each frame in Fig.2. According to the poses, the simulator will generate GS images, then synthesize RS images from these GS images by extracting the corresponding row pixels of the corresponding GS image. In experiments, we set the image resolution as $900\times900$ with a 1384.6 focal length, the magnitude of $\mathbf{d}$ is 0.3, and the magnitude of $\mathbf{w}$ is fixed to $0.4\degree$. Besides, both translation and rotation include motion on x, y, and z-axis. Then we use SIFT \cite{Lowe_SIFT_2004} to extract feature points and RANSAC to reduce the effect of noise on pose estimation. Our experimental result is reported in Fig.(4).
Fig.4(a) shows the original RS image. And in Fig.4(c).(f), we can observe the depth truth of RS image and the depth error between depth truth and depth estimated by the SGM algorithm. We notice that the left part of the depth error map is set to 0 because the SGM algorithm cannot estimate the depth value of the left area on the left image.
Fig.4(b) shows the corrected image using our method and the depth estimated by the SGM algorithm. 
In Fig.4(d), we overlay the original RS image and the ground truth GS image rendered by the pose of the first row, the red area shows the distortion. And then we overlap the corrected image and the GS image as shown in Fig.4(e). Comparing Fig.4(d) and Fig.4(e), we can find that the rectified RS image almost coincides with the GS image, except for the surrounding area where the SGM algorithm cannot estimate the depth and a small area where the depth error is more than 2m. 
The experiments prove that our proposed algorithm can also estimate the motion well, and the SGM algorithm can be used to estimate the depth of stereo RS images and get an accurate depth value which can be used in RS image correction.

\vspace{-0.5cm}
\section{CONCLUSION}
\label{sec:typestyle}
In this paper, we have proposed a novel linear relative pose solver for stereo rolling shutter cameras under consecutive frames using 9 pairs of correspondences for the left and right camera respectively. We showed that for RS stereo images, traditional SGM algorithm can be used to estimate the depth map and achieve an accurate result. Experimental results on synthetic RS images show that we can well correct the RS image using our proposed relative pose solver and depth estimated by the SGM algorithm.

\noindent\textbf{Acknowledgements:} This research was supported in part by the National Natural Science Foundation of China under Grants 61871325, 61420106007, and 61671387 and the National Key Research and Development Program of China under Grant 2018AAA0102803.


\bibliographystyle{IEEE}
\bibliography{reference}

\end{document}